\documentclass[10pt,sigconf,acmlarge,screen,nonacm]{acmart}
\usepackage{setspace}
\usepackage{booktabs}
\makeatletter
\def\@copyrightspace{\relax}
\let\@authorsaddresses\@empty
\begin{document}

\title {Generative Modeling of Networked Time-Series via Transformer Architectures}
\raggedbottom
\author{Yusuf Elnady}
\email{yelnady@vt.edu}
\affiliation{%
  \institution{Virginia Tech}
  \postcode{24060}
  \country{USA}
}
\begin{abstract}
\textbf{Abstract}\\ 
  Many security and network applications require having large datasets to train the machine learning models. Limited data access is a well-known problem in the security domain. Recent studies have shown the potential of Transformer models to enlarge the size of data by synthesizing new samples, but the synthesized samples don't improve the models over the real data. To address this issue, we design an efficient transformer-based model as a generative framework to generate time-series data, that can be used to boost the performance of existing and new ML workflows. Our new transformer model achieves the SOTA results. We style our model to be generalizable and work across different datasets, and produce high-quality samples.
\end{abstract}


\maketitle

\section{Introduction}
Limited data is a large issue in the security and networks community. It is hard to obtain data for new bots or new attacks, as there can be only just a few samples that represent this new class. Limited access to data is a long-term obstacle to science powered by data. In networking and systems analysis, data-driven techniques are important since network providers and infrastructure programmers use data to make new data-driven management decisions. Since machine learning and deep learning are eager for large datasets, it becomes a problem to train machine learning models to work best when we don't have enough data points from these new classes. This imbalanced class representation remains an issue in many domains, although there are many ways to increase the size of such datasets. Some of the ways include the augmentation or up-sampling techniques, but in security datasets, these methods are not used too much. Therefore, we want to have a method that, given a sample, it can generate many samples that are similar and representative of the original data. Hence, we can have big data that would be useful in enhancing and retraining our ML models to defend against the new attacks. We also want these generated samples to be done automatically without any human expertise, so they can generalize to all kinds of network and systems downstream tasks. Our work focuses on a generalized version of datasets, where each data point consists of time-series measurements at different time-steps and is associated with metadata that represents the properties of this data point. 

\section{Problem Statement}
The goal of this project is to explore the use of ML-generated synthetic data to improve ML tasks in network security/networks. This includes the use of generative models like GANs and state-of-the-art models, i.e., transformers. We want to synthesize time-series data, that can be utilized to boost the performance of existing and new ML workflows. Furthermore, we want the model to improve resilience against adaptive attackers and to style it as a generalizable model that can work across different datasets and produce high-quality and high-fidelity samples. Although the ultimate goal would be the downstream tasks, we also focus on the structural characterization of the generated datasets to assure that it captures the temporal correlation similar to the real data. We consider many downstream tasks that our generated data can be applied to it to increase the efficiency, accuracy, and robustness of existing ML workflows. Downstream tasks include network intrusion detection, malicious user detection (e.g., social media), bot detection, malware detection (based on sequential API traces), malicious traffic classification (e.g., DNS traffic for bot communication), and IoT security classifiers.
  
\section{Background \& Related Work}
Generative models are the core of this project, and there are many generative frameworks such as Autoregressive models, Markov Models, Deep Belief Networks (DBN), Variational Auto Encoders (VAE) \cite{VAE}, and Generative Adversarial Networks (GANs) \cite{GAN}. GANs are unsupervised models and data-driven generative modeling techniques \cite{GAN} that learn the distribution of the data from the training samples, then start generating samples similar to the real data. Therefore, generated samples are drawn i.i.d. from the original distribution. GANs are mainly designed for the computer vision domain, but they can generalize to other domains by replacing the convolutional neural networks \cite{CNN} inside the generator and discriminator of the GANs.
Although GANs have proved to work very well in many fields, RNN GANs suffer from many issues in time-series real-valued data \cite{esteban2017realvalued}: it captures temporal correlations poorly, it cannot correlate the generation of complex multidimensional relationships (i.e., time-series measurements and the associated metadata), and it generates only a few modes of the underlying distribution (i.e., mode collapse) \cite{wasserstein}. The generator in GANs typically takes as inputs prior measurements and Gaussian noise and outputs one measurement at a time. The two most relevant papers to our model are TimeGAN \cite{timegan} and  DoppelGANger \cite{doppelGANger}, but mine only depends on the transformer models.
\subsection{TimeGAN} RNNs are used in both the generator and discriminator in TimeGAN \cite{timegan}. TimeGAN divides datasets into smaller time series, each with 24 epochs, and only evaluates the model when new time series of this length are produced. This isn't feasible in the networks and systems domain because important properties of data are always observed over longer time scales, and the data tends to have very long sequences. TimeGAN trains an additional neural network that maps time series to vector embeddings, and the generator outputs sequences of embeddings rather than samples, which can be transformed back to their original representation. TimeGAN cannot jointly create high-dimensional metadata and time series of different lengths, and in the area of security and networks, the sample lengths are not fixed and depend only on when the measurements have been done and when the measurements have stopped.
\subsection{DoppelGANger}
DoppelGANger \cite{doppelGANger} is the current state-of-the-art, outperforming TimeGAN. It is a custom workflow for data sharing that tackles many challenges faced by TimeGAN. DoppelGANger is a computationally expensive and very complicated model that addresses the issues of capturing long-term effects, tackling mode collapse \cite{wasserstein}, and capturing attribute relationships. These challenges are solved using many tricks. It models the correlations between measurements and their metadata by decoupling the generation of metadata from time series and feeds metadata to the time series generator at each time step, and introduces an auxiliary discriminator for the metadata generation. The architecture doesn't suffer from mode collapse \cite{wasserstein} because it separately generates randomized max and min limits and a normalized time series, which can then be rescaled back to the realistic range. IT captures temporal correlations by using batched samples rather than singletons. However, the ultimate goal is the final tasks (aka. downstream tasks), and our transformer model is much simpler and can generate samples with much higher fidelity than DoppelGANger. Moreover, our model is superior and achieves higher accuracy and fidelity than the DoppelGANger architecture.

\section{Model Structure}
\subsection{Transformer Background}
The original transformer architecture is proposed in the paper "Attention is all you need" \cite{attention}. The vanilla transformer is designed as a seq2seq problem for neural machine translation (NMT) tasks in the Natural Language Processing (NLP) domain. Transformers have proved to be the SOTA model for seq2seq problems, and recently, it has been applied to different domains such as the computer vision field. Transformers are better than RNNs. They totally avoid recursion and can be parallelized (training is faster, and they can learn long-term temporal correlation relationships between words for very long sentences. Transformer models apply a lot of tricks: Positional Encoding or Embedding, Self-Attention, Multi-Head Attention, Masked Attention, Residual Connections, Layer Norm, Positional Feed-Forward Network, No RNN cells at all \cite{attention}. The transformer consists of Encoder and Decoder Blocks. In a typical NMT task from English to French, the first encoder block receives the English sentence, and the first decoder block receives the French sentence. But in the Natural Language Generation task, we only have the English sentence, which is our sample, and we want the model to learn how to reproduce it again and forecast the next timesteps. That is, the targets are just the inputs shifted to the left. The challenge is applying transformer models on real-valued (time-series) data.

\subsection{The New Architecture}
The original transformer described above cannot be applied directly to real-valued data, which is the dominant type of dataset in security and networks. We modify the vanilla transformer and enhance it to generalize it to time-series tasks. We only use the encoder blocks of the transformer, and we ignore the decoder blocks all. \\The encoder blocks of the original transformer consist of Self Attention and Positional Feed-Forward Network, but we further improve the encoder blocks and use an architecture similar to GPT2 \cite{gpt2}. Our encoder blocks use two masks. The first mask is the Masked-Self Attention to prevent each time step from attending to future samples, and only pays attention to the previous ones. The second mask is the Padding Mask because in many datasets, we have different samples of different lengths of timesteps that are all padded to a maximum length. Therefore, we don't want the multi-head attention to consider the padding values. \\The input embedding layer is replaced by a single-layer neural network that maps the data dimensions from the number of features to the hidden dimension that is expected by the encoder blocks. As transformers don't use any RNN cells, they need a way to identify the positions of timesteps, as a result, we use the original positional encoding \cite{attention}. Although there are other options, such as time2vec and positional embedding, positional encoding seems to provide acceptable results. 
Softmax activation of the output layer is replaced by a Sigmoid or Tanh activation function based on the data normalization range. The softmax layer is removed because we want to produce real-valued numbers, not probabilities, as we don't have a vocabulary or corpus in our case. The overall architecture is shown in Figure~\ref {fig:transformer}.
\subsection{Pipeline} 
\subsubsection{The Training Phase}
\begin{figure}[h!]
  \includegraphics[scale=0.28]{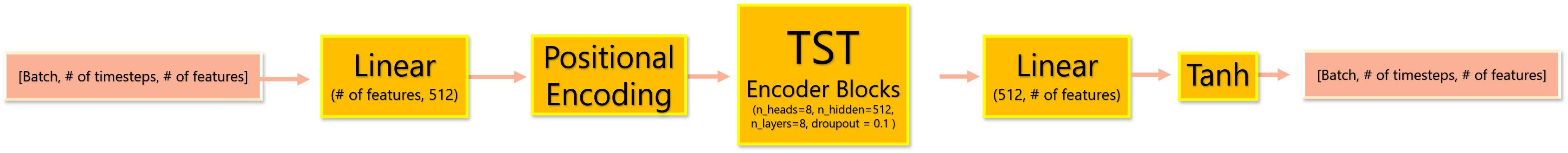}
  \caption{Our New Transformer Architecture}
  \label{fig:transformer}
\end{figure}

\hfill\\Given the architecture in figure ~\ref{fig:transformer}, we compose the samples into batches of this format [Batch Size, \# of timesteps, \# of features], then they are fed into our transformer model. The current architecture has 8 heads in the attention layers, 8 encoder blocks, hidden dimension of 512 for the Q, V, and K matrices. We also add a 0.1 dropout. The loss function is MSE loss, and we use a custom masked version of MSE loss to calculates the loss only for the actual timesteps and ignore the paddings. The batch size is 64, and the window size is 400 timesteps (non-sliding). Note that all of these hyperparameters can be tweaked, but due to memory limitations, we chose these values. The number of training epochs is 400. The number of parameters in the transformer model is 12,635,659. The optimizer is Adam with a learning rate of 0.0001. The targets are the same as the input, except that they are shifted one step to the left.  
\subsubsection{The Generation Phase}
\hfill\\ We use the trained transformer for generating the synthesized dataset. The current transformer architecture is conditional and requires a seed at the beginning, so we seed two timesteps at the start of the transformer, and it can generate the next timesteps in an auto-regressive way. Note that the generation/decoding phase for one sample cannot be parallelized. To stop the generation at the correct timestep, we need the model to learn the sequence lengths of timesteps as part of the training. The encoder blocks again employ the masked self-head attention so that the forecasted measurements can be obtained. depend only on previously seen measurements. We use the method of generation flags discussed in this paper \cite{doppelGANger} to add two features for each timestep. That should allow the model to halt at the correct timestep. Note that we cannot borrow the stop tokens options from the NLP domain since we do not have a corpus, but we do have real values for our datasets. We generate samples from any class we want by feeding the seed from any data point that belongs to this class. The model described is generalizable, can expand across various datasets and use cases, and achieve high accuracy.
\subsubsection{Downstream Tasks}
\begin{figure}[h!]
  \includegraphics[scale=0.35]{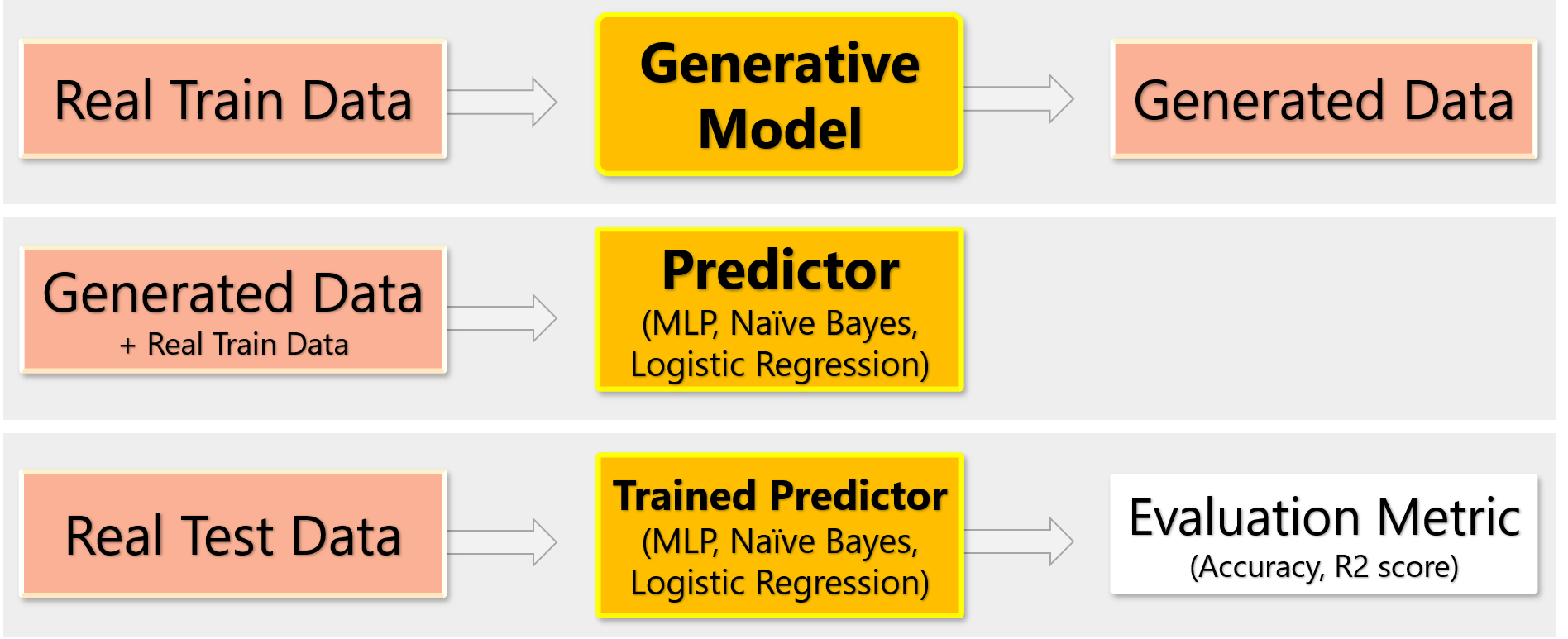}
  \caption{The Pipeline: 1- Training 2- Generating 3- Downstream Tasks }
  \label{fig:pipeline}
\end{figure}

The downstream task can be a classification or regression/forecasting task. We train the ML models of the downstream tasks on the synthesized dataset from the generator. These ML models can be MLP, Naïve Bayes, Logistic Regression, SVM, etc. We also train the ML models on the synthesized dataset + a proportion of the real dataset to investigate how our samples perform in case of a limited few samples from a specific class, which is the case if there's a new bot or attack. The test is only performed on real data to evaluate the generated dataset in the real world. Figure ~\ref{fig:pipeline} summarizes the general pipeline.

\section{Datasets}

We compare our model against DoppelGANger, which is the SOTA, on two datasets, and we plan to do more investigations on more datasets in the future. The two datasets are Google Cluster Usage Traces \cite{GCUT} and Wikipedia Web Traffic \cite{WWT}. Both datasets contain 50,000 real data points, and we split each of them in a 50:50 ratio into real train datasets and real test datasets.
\subsection{Google Cluster Usage Traces (GCUT)}
\begin{figure}[h!]
  \includegraphics[scale=0.4]{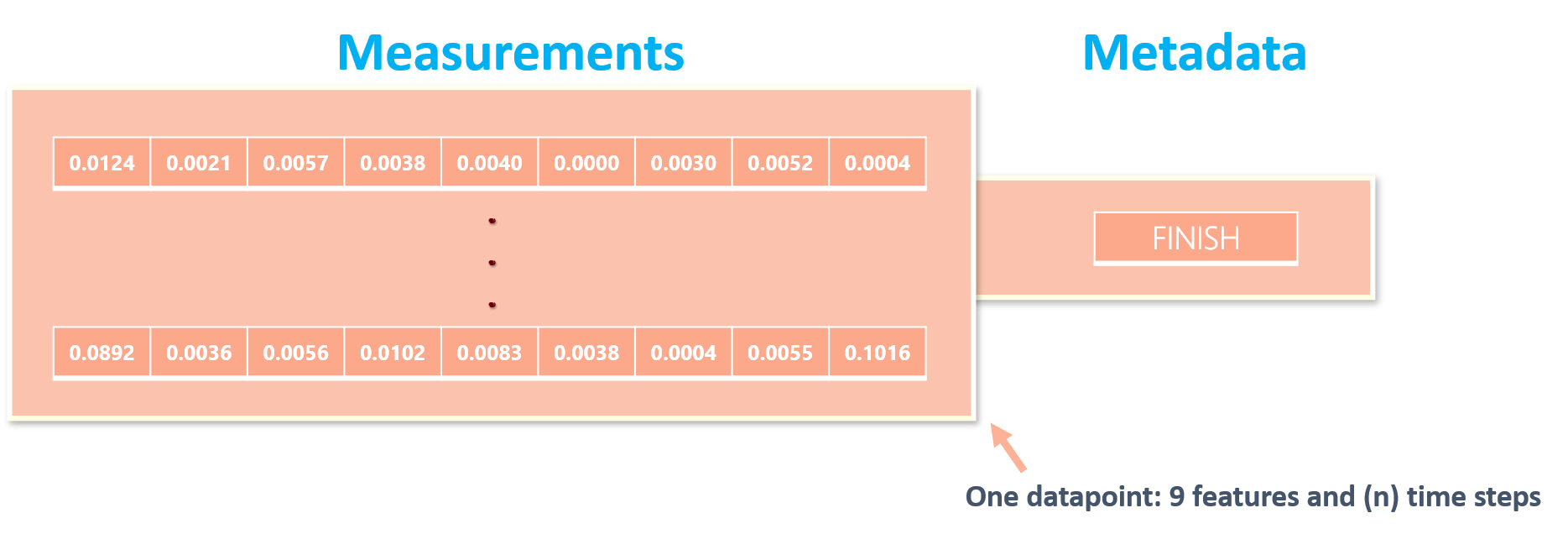}
  \caption{Architecture of one datapoint in the GCUT dataset}
  \label{fig:typical}
\end{figure}

\begin{table}[]
\begin{tabular}{|c|c|}
\hline
\textbf{Measurements}  & \textbf{Metadata/Attributes} \\ \hline
CPU Rate               & End Event Type               \\ \hline
Canonical Memory Usage &                              \\ \hline
Assigned Memory Usage  &                              \\ \hline
Unmapped Page Cache    &                              \\ \hline
Total Page Cache       &                              \\ \hline
Maximum Memory Usage   &                              \\ \hline
Local Disk Space Usage &                              \\ \hline
Maximum CPU Rate       &                              \\ \hline
Sampled CPU Usage      &                              \\ \hline
\end{tabular}
\caption{\label{tab:gcut}Scheme of GCUT Dataset.}
\end{table}
Google cluster is a set of machines, packed into racks, and connected by a high-bandwidth cluster network \cite{GCUT}. This dataset is the  v2.1 trace from 2011 and contains usage traces of a Google Cluster of 12.5k machines over 29 days in May 2011. Due to the huge size of this dataset and the substantial memory requirements to train transformer models, we uniformly sample a subset of 100,000 tasks and use their corresponding measurement records to form the dataset \cite{doppelGANger}. This dataset is composed of features and attributes that are shown in Table~\ref {tab:gcut}. Each task consists of a variable number of time-steps, where each timestep describes the nine measurements captured at this timestep, and the overall task is associated with one metadata that is the end event type of this task. The end event type can be FAIL, FINISH, KILL, or EVICT. The average length of this dataset is 11 timesteps, and the maximum length is 2500 timesteps. A typical data point is shown in Figure ~\ref{fig:typical}.

\subsection{Wikipedia Web Traffic (WWT)}

\begin{table}[]
\begin{tabular}{|c|c|}
\hline
\textbf{Measurements} & \textbf{Metadata/Attributes} \\ \hline
Number of Views       & Wikipedia Domain             \\ \hline
                      & Access Type                  \\ \hline
                      & Agent Type                   \\ \hline
\end{tabular}
\caption{\label{tab:wwt}Scheme of WWT Dataset.}
\end{table}

This dataset consists of approximately 145k time series \cite{WWT}. Each of these time series represents a number of daily views of a different Wikipedia article, starting from July, 1st, 2015 up until December 31st, 2016. All samples have a measurement length of 550. The timesteps are 1-d features, which contain the number of daily views for one Wikipedia page. The features and attributes of WWT are shown in the table ~\ref{tab:wwt}.

\section{Experiments \& Results}
\subsection{Evaluation Metrics}
Evaluation metrics are important to analyze the performance of our models, and we consider a different set of evaluation metrics depending on the type of the downstream task. For example, in classification problems, we may use the accuracy and F-Score (class-wise) as the evaluation criteria. F-score is the harmonic of the precision and recall that is $$
F_1 = 2*\frac{Precision * Recall}{Precision + Recall}$$ \\In regression tasks, we focus on the R2 score (coefficient of determination) which provides a measure of how well-observed outcomes are replicated by the model, based on the proportion of total variation of outcomes explained by the model; Higher is the better. R2 score is $$R^2 = 1-\frac{SS_{res}}{SS_{tot}}$$ that is one minus (The sum of squares of residuals divided by the total sum of squares). \\We also emphasize that the generated samples have the same structural characterization as the real samples to preserve the trends and distributions well enough to reveal such structural insights. We use the Pearson correlation to calculate and compare the cross-measurement correlation between any two measurements/features. Pearson correlation coefficient (PCC) is a measure of linear correlation between two sets of data and takes ranges of values from +1 to -1. PCC is defined as $$\rho_{X, Y} = \frac{cov(X, Y)}{\sigma_X\sigma_Y}$$ A value of 0 indicates that there is no association between the two variables, and a value greater than 0 indicates a positive association; that is, as the value of one variable increases, so does the value of the other variable \cite{pearson}. 
\subsection{GCUT Results - Classification}
\begin{figure}[h!]
  \includegraphics[scale=0.48]{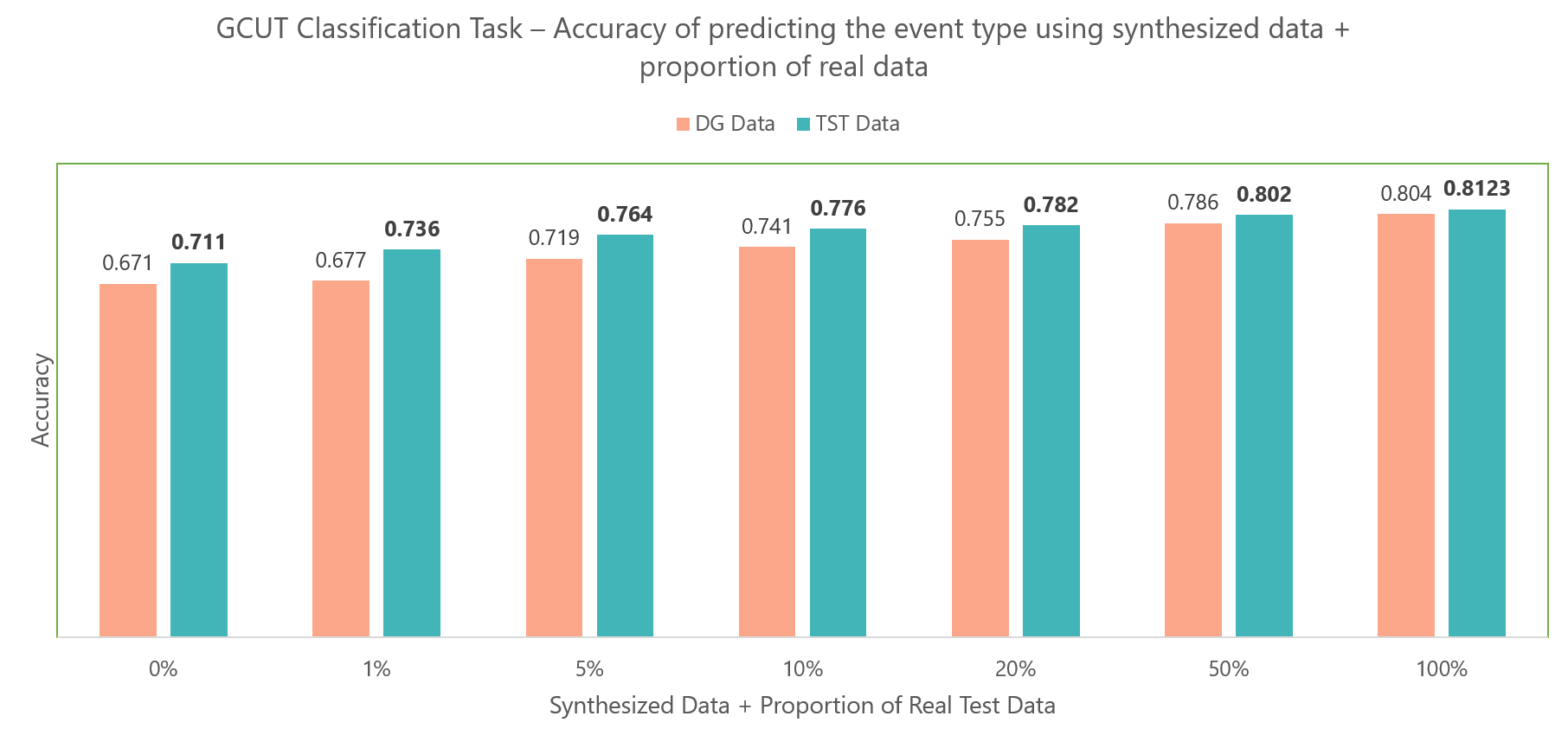}
  \caption{Accuracy of predicting the event type using synthesized data + proportion of real data}
  \label{fig:gcut_acc}
\end{figure}

\begin{figure}[h!]
  \includegraphics[scale=0.45]{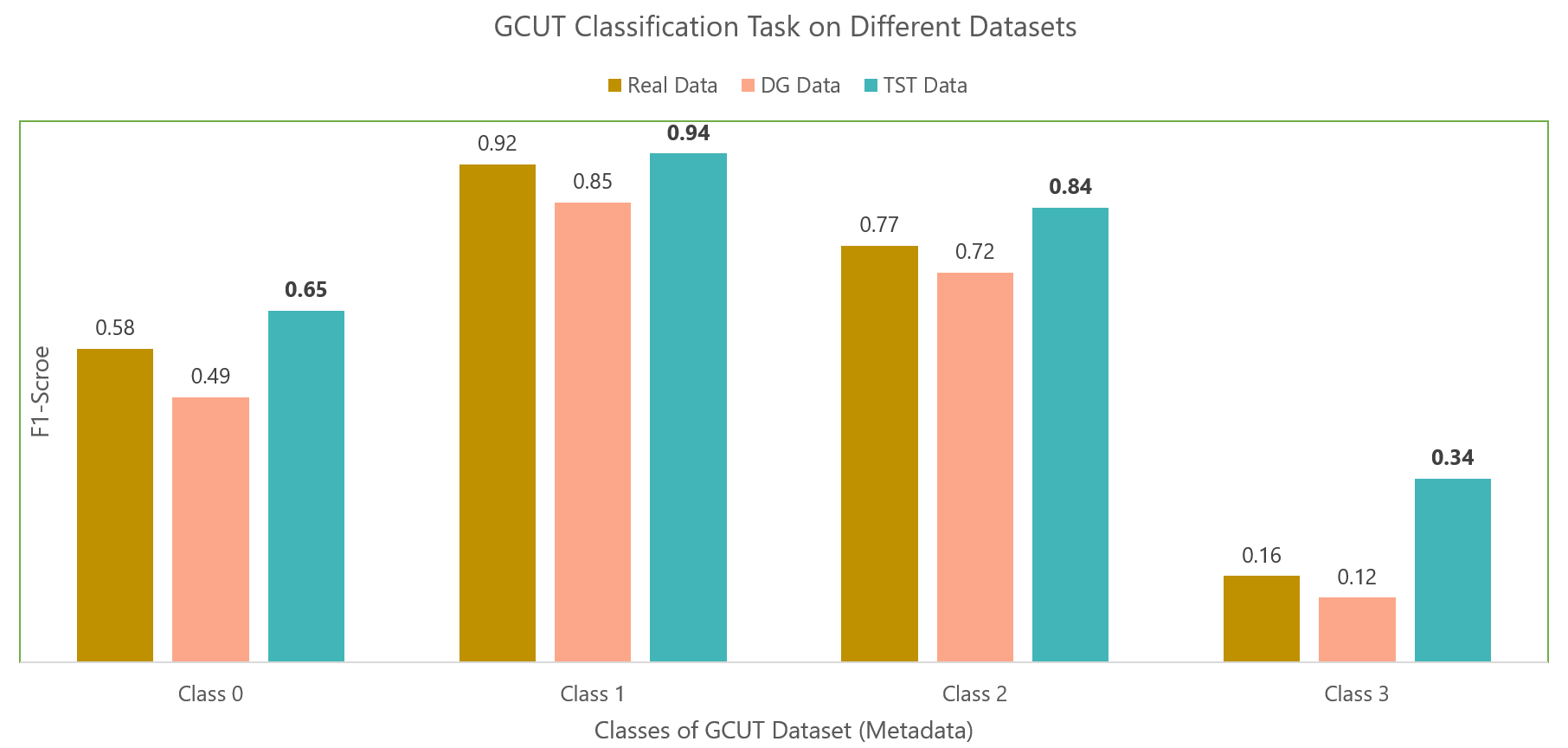}
  \caption{F-score of GCUT Classification Task on Different Datasets}
  \label{fig:gcut_f}
\end{figure}

In GCUT, the system measures its resource usage. Once a task finishes, it assigns an end event type for itself. The end event type has 4 classes: FAIL, FINISH, KILL, and EVICT. Our goal is to classify the task based on its 9 measurements. We train a one-layer neural network for 100 epochs on generated data + different proportions of the real train data, and show the results for different proportions. The evaluation metrics are always tested on the real test data. We compare our model (TST) results against DoppelGANger results. Figure ~\ref{fig:gcut_acc} shows the accuracy of our model (TST) and DoppelGANger, and we beat the accuracy of DoppelGANger in all of the settings. Figure ~\ref{fig:gcut_f} shows the F-score a class-wise for TST, DoppelGANger, and real datasets. 
\subsection{WWT Results - Regression}
The WWT dataset consists of one measurement, which is the number of daily views for one Wikipedia page. The regression task is: given the first 500 consecutive days page count views of this Wikipedia article, predict the page count views for the next 50 days. We compare our results against 4 models: one-layer neural network, five-layer neural network, Linear Regression, and Kernel Ridge. Table ~\ref{tab:wwt} shows the R2 score of different datasets on the 4 regressors. Note that our (TST) dataset + real dataset achieves the highest R2 score for the first two models. 

\begin{table}[]
\begin{tabular}{|l|l|l|l|l|}
\hline
\textbf{Dataset/Regressor Model} & \textbf{MLP  1-layer} & MLP 5-layers    & Linear Regression & Kernel Ridge    \\ \hline
Real Data                        & 0.9061                & 0.8941          & \textbf{0.9088}   & \textbf{0.9087} \\ \hline
TST Data + Real Data             & \textbf{0.9093}       & \textbf{0.9034} & 0.9073            & 0.9070          \\ \hline
DG Data + Real Data              & 0.8916                & 0.8909          & 0.9007            & 0.9070          \\ \hline
\end{tabular}
\caption{\label{tab:wwt}R2 Score on WWT Dataset regression task.}
\end{table}
\section{Challenges \& Future Work}
The current generative framework can be further improved in many directions.
\begin{itemize}
    \item Training the model to generate variable sequence length samples (i.e., generating sequences of the correct length automatically). Currently, the model doesn't capture the lengths very well, and we investigate different tricks of adding some features (or magic rows) to each timestep of the data points during the training process.
    \item Enabling the model to work in an unconditional generation setting where no seed (initial timesteps) is required to start the decoding or the generation phase.
    \item Adding the controlled generation option to allow generating samples based on specified metadata. For example, by asking the transformer to generate some samples from class 0, we will get new synthesized samples that fall in the same cluster of class 0. 
    \item Improving the controlled generation to be conditioned on an existing sample. This will be useful to adapt against an adaptive attacker.
    \item Using more advanced transformer encoder blocks. The current model uses the same attention layer from the original transformer paper. We want to experiment with much newer attention layers, such as the Informer model \cite{informer} that uses lower computing resources, trains faster, and solves memory issues for long sequence time-series forecasting. 
    \item Compare results on various other downstream tasks.
\end{itemize}

\section{Conclusion}

In this project, we studied the limitations of the current generative models. We designed a new generative framework that generates synthetic time-series data to boost the performance of existing and new machine learning workflows. We compared our results against DoppelGANger on two datasets and proved that our model achieves higher fidelity and improves the accuracy on the downstream tasks. 



\bibliographystyle{ACM-Reference-Format}
\bibliography{TST}

\end{document}